\documentclass{article} 
\usepackage{iclr2027_conference,times}
\iclrfinalcopy

\usepackage{amsmath,amsfonts,bm}

\def\eqref#1{equation~\ref{#1}}

\def\1{\bm{1}}

\DeclareMathAlphabet{\mathsfit}{\encodingdefault}{\sfdefault}{m}{sl}
\SetMathAlphabet{\mathsfit}{bold}{\encodingdefault}{\sfdefault}{bx}{n}

\usepackage[hidelinks]{hyperref}
\usepackage{url}
\usepackage{booktabs}
\usepackage{graphicx}
\usepackage{multirow}
\usepackage{pifont}

\usepackage{amsmath, amssymb}
\usepackage{tabularx}
\usepackage{booktabs}
\usepackage[skins]{tcolorbox}
\usepackage{centernot}
\usepackage{twemojis}
\usepackage{pifont}
\usepackage{xspace}
\usepackage[table, dvipsnames]{xcolor}
\usepackage{booktabs}
\usepackage{wrapfig}
\usepackage{comment} 
\usepackage[nameinlink]{cleveref}

\usepackage{hyperref}
\hypersetup{
    colorlinks=true,
    linkcolor=Blue,      
    citecolor=Blue,      
    urlcolor=Blue,    
    filecolor=Black
}

\newcommand{\effy}[1]{\textcolor{red}{[Effy: #1]}}

\newcommand{\madelon}[1]{\textcolor{violet}{[Madelon: #1]}}

\newcommand{\daniel}[1]{\textcolor{olive}{[Daniel: #1]}}

\newcommand{\as}[1]{\textcolor{teal}{[Aécio: #1]}}

\newcommand{\header}[1]{\vspace{1mm}\noindent\textbf{#1}.}

\newcommand{\qwen}[2]{\texttt{qwen-#1-#2\xspace}}
\newcommand{\qwenSmall}[0]{\qwen{3.5}{4b}\xspace}
\newcommand{\qwenMedium}[0]{\qwen{3.8}{27b}\xspace}
\newcommand{\geminiFlash}[1]{\texttt{gemini-#1-flash\xspace}}

\newcommand{\gptSol}[0]{\texttt{gpt-5.6-sol}\xspace}

\newcommand{\datasetlinks}{%
\footnote{\tiny
Kaggle: \href{https://www.kaggle.com/}{kaggle.com}
\quad
OpenML: \href{https://www.openml.org/}{openml.org}
\quad
UCI: \href{https://archive.ics.uci.edu/}{archive.ics.uci.edu}}}

\definecolor{redbg}{HTML}{F8D7DA}
\definecolor{redborder}{HTML}{F5C2C7}
\definecolor{redtext}{HTML}{842029}

\definecolor{bluebg}{HTML}{CFE2FF}
\definecolor{blueborder}{HTML}{B6D4FE}
\definecolor{bluetext}{HTML}{084298}

\definecolor{greenbg}{HTML}{D1E7DD}
\definecolor{greenborder}{HTML}{BADBCC}
\definecolor{greentext}{HTML}{0F5132}

\newtcbox{\leakbadge}{
  on line, arc=2pt, outer arc=2pt,
  colback=redbg, colframe=redborder,
  colupper=redtext,
  boxsep=0.25pt, left=2pt, right=2pt, top=1pt, bottom=1pt,
  boxrule=0.4pt,
  fontupper=\tiny\sffamily\bfseries
}

\newtcbox{\otherbadge}{
  on line, arc=2pt, outer arc=2pt,
  colback=bluebg, colframe=blueborder,
  colupper=bluetext,
  boxsep=0.25pt, left=2pt, right=2pt, top=1pt, bottom=1pt,
  boxrule=0.4pt,
  fontupper=\tiny\sffamily\bfseries
}

\newtcbox{\tfbadge}{
  on line, arc=2pt, outer arc=2pt,
  colback=greenbg, colframe=greenborder,
  colupper=greentext,
  boxsep=0.25pt, left=2pt, right=2pt, top=1pt, bottom=1pt,
  boxrule=0.4pt,
  fontupper=\tiny\sffamily\bfseries
}

\newcommand{\llmjudge}{%
  \twemoji[height=1.0em]{balance scale}%
}
\newcommand{\testsuite}{%
  \twemoji[height=1.0em]{toolbox}%
}
\newcommand{\humansolution}{%
  \twemoji[height=1.0em]{magnifying glass tilted right}%
}

\newcommand{\stage}[1]{%
  \begingroup
  \setlength{\fboxsep}{1.8pt}
  \colorbox{black!6}{\footnotesize\ttfamily\detokenize{#1}}%
  \endgroup
}

\newtcolorbox{myframe}[2][]{%
 enhanced,colback=white,colframe=black,coltitle=black,
  rounded corners=north,
  boxrule=0.4pt,
  top=4pt,
  left=4pt,
  bottom=3pt,
  right=4pt,
  attach boxed title to top left={yshift=-0.3\baselineskip-0.4pt,xshift=2mm},
  boxed title style={tile,size=minimal,left=0.5mm,right=0.5mm,
    colback=white,before upper=\strut},
  title=#2,#1
}

\title{The Silent Failures in Agentic Data Science}

\author{Xue Li$^1$, Hongqian Xia$^1$, Daniel Gomm$^1$, Aécio Santos$^1$, Cornelius Wolff$^1$, Madelon Hulsebos$^1$~\\
\\
$^1$Centrum Wiskunde \& Informatica, Amsterdam, Netherlands\\
\texttt{\{xuel, madelon\}@cwi.nl}
}

\begin{document}

\maketitle

\begin{abstract}

\vspace{-0.2cm}
Large language models have significantly lowered the barrier to writing code. Naturally, this has also led to the development of data science agents that promise automating the design and execution of data science from task formulation to model development. Existing evaluations of agentic data science (agentic DS) assess downstream predictive performance and code correctness, providing preliminary insight in agentic DS capabilities. We argue that these evaluations overlook the fundamental principles of data science: methodological data science- and domain knowledge. As such, only assessing downstream performance masks methodological failures such as data leakage and inappropriate validation, limiting the development of robust agentic DS systems. In this paper, we evaluate agentic DS \textit{beyond accuracy} from three different angles: 1) a large-scale analysis of failure modes in common agentic DS frameworks and models across 220 data science tasks, 2) a reusable and extensible test suite with curated data science tasks that test for six specific failures in a controlled manner, and 3) a qualitative assessment of agentic DS pipelines against expert solutions on Kaggle competitions. 
We find that the majority of agentic solutions contain at least one failure mode despite the complexity of agentic harness or model capacity. Moreover, we identify a positive correlation between the occurrence of data leakage failures and predictive performance, confirming the necessity to evaluate agents \textit{beyond accuracy}. Our qualitative analysis shows that DS agents generally lack problem solving creativity and domain knowledge, compared to competition-winning data scientists. We believe that principled failure mode analysis, as presented here, should be integral part of agentic DS evaluations, and that the insights we surface will be key in developing trustworthy agentic DS pipelines. Project resources are available at \url{https://trl-lab.github.io/ds-agent-failures/}.

\end{abstract}

\section{Introduction}
\vspace{-0.2cm}
The strong code generation capabilities in large language models (LLMs) have led to their wide adoption in software engineering~\citep{jimenez2024swe-bench}. When connected to execution environments, LLMs can be deployed as agents for the autonomous completion of tasks, which has become an emerging practice across applications and industries~\citep{pan2025agents-production}. While coding agents generally improve with more capable backbone LLMs~\citep{jimenez2024swe-bench}, understanding performance beyond end-to-end capabilities is key~\citep{gu2025aiswe-challenges}. Prior analyses of agentic code-based systems have surfaced common failure modes beyond code correctness, such as disobeying task specifications, misaligned agent interactions, and incorrect task verification~\citep{cemri_why_2026}. 

In parallel with coding agents, we observe an influx in the automation of data science pipelines through specialized \textit{data science agents} as well~\citep{guo_dsagent_2024a, grosnit_kolbbased_2025, du_mlevolve_2026}. Data science use cases, particularly of a predictive nature, are traditionally characterized as multi-disciplinary processes that involve long-trained data scientists, domain experts, software and data engineers~\citep{debie2022automatingds}. What constitutes a ``good'' data science pipeline is also inherently task- and data-dependent, and cannot be coupled with a hard verifiable signal. For example, model training and evaluation may require a temporal splitting procedure if the use-case demands temporal anticipation, while predictive performance metrics need to suit the target data distribution. These characteristics also translate into different evaluation protocols: whereas software is tested in an integrated verifiable manner, the assessment of (statistical) validity and utility in data science is typically spread over the entire process and requires input from domain experts, engineers, and data scientists~\citep{debie2022automatingds, cong2025position}. These characteristics, therefore, necessitate different evaluations beyond code generation capabilities.

An early large-scale evaluation of generic agents across diverse tasks common in industry, among which data science tasks, highlighted a significant difference where most LLMs had a success rate of zero on data science tasks compared to approximately 15\% higher success rates on other tasks~\citep{xu2026theagentcompany}. Specialized DS agent benchmarks provide a more holistic analysis of metrics such as cost, efficiency, and success rates, surfacing the differences across models and DS agent frameworks~\citep{huang_mlagentbench_2024, jing_dsbench_2024, nayak_benchmarking_2026}. However, \textit{we argue that the potential presence of silent failures can give a distorted view of generalizability, hence making it risky to assess the actual capability of DS agents with downstream metrics only.} The development of reliable DS agents, therefore, demands more fine-grained evaluations beyond downstream performance metrics to understand their true failure modes and capabilities.

\begin{myframe}[colframe=blue, title=\textcolor{blue}{\textbf{Silent Failures}}]

We define \emph{silent failures} as task-relevant methodological violations that
compromise the validity or soundness of a solution, but may remain undetected by conventional output-based evaluation. Silent failures can coexist with executable code and strong predictive performance.

\end{myframe}

\begin{wrapfigure}{r}{0.4\textwidth}
    \vspace{-0.3cm}
    \centering
    \includegraphics[width=\linewidth]{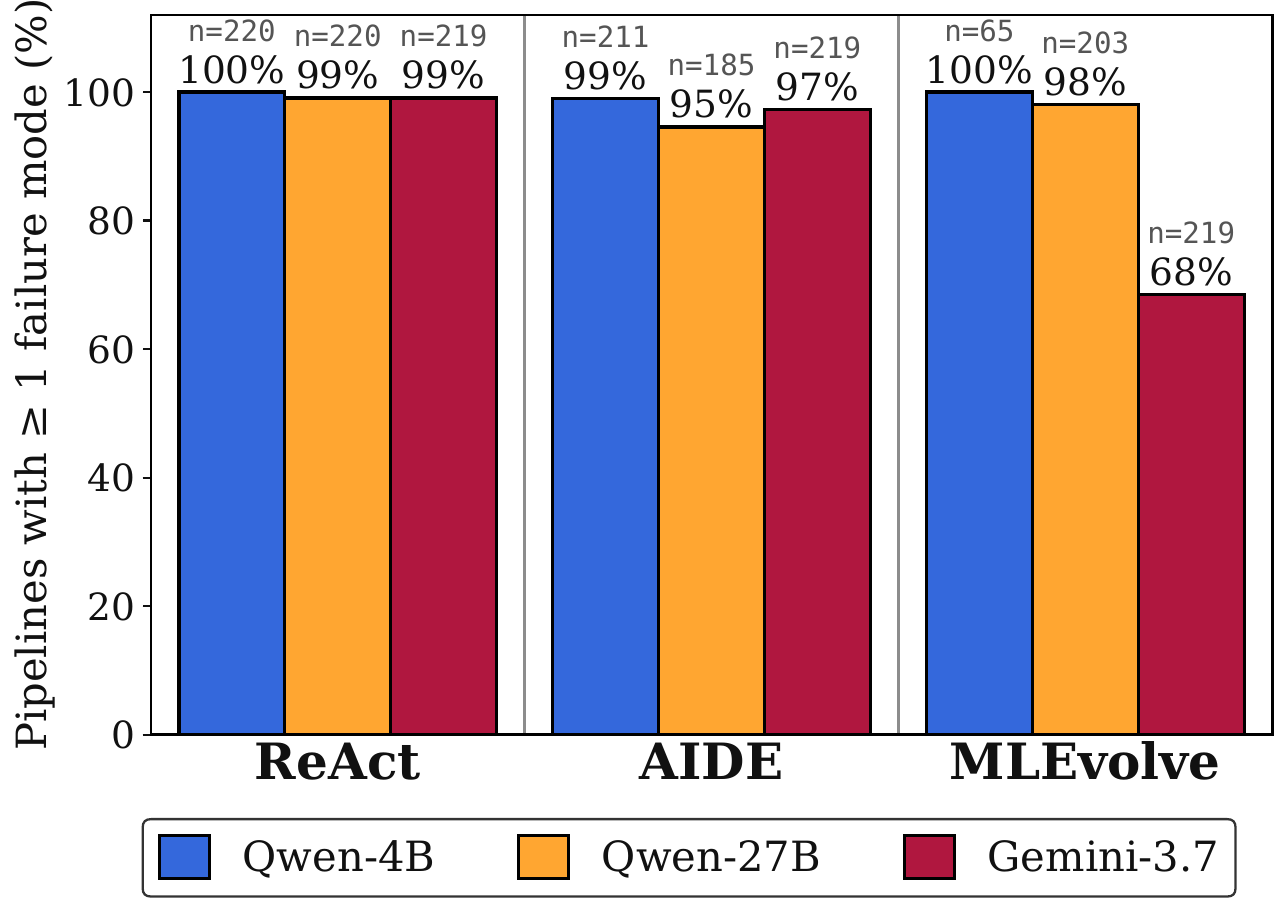}
    \vspace{-1em} 
    \caption{Percentage of pipelines with at least one failure mode per agent-model configuration.
    }
    \label{fig:pct_pp_agent}
    \vspace{-.7em} 
\end{wrapfigure}

In this paper, we present \ding{172} the first large-scale analysis of common failure modes of various DS agents and models on 220 tasks, \ding{173} insights from a reusable controlled test suite with curated tasks, and \ding{174} a qualitative analysis of capabilities in comparison with human expert solutions, all grounded in data science best practices~\citep{lones2024avoidingpitfalls,kumar2025navigatingpitfalls}.
Across analyses, we find that the majority of agent-generated pipelines contain at least 1 data science failure mode regardless of the complexity data science agent framework (Figure~\ref{fig:pct_pp_agent}). We also find that data leakage failure rates correlate with inflated predictive performance, underlining the necessity to evaluate agents beyond accuracy. Finally, our qualitative analysis shows that agents’ solutions are weaker on task formulation and feature engineering compared with human expert solutions, illustrating a lack of problem-solving creativity and domain knowledge. We believe that principled failure mode analysis, as presented in this paper, should be an integral part of agentic DS benchmarks and evaluations. We also envision that the DS practices framework and surfaced insights are key for developing robust data science agents.

\vspace{-0.15cm}
\section{Related Work}
\vspace{-0.25cm}
\header{Automated data science} Automated Machine Learning (AutoML) systems address the need for automating the selection of model architectures and optimization of hyperparameters~\citep{hutter_automated_2019}
, through frameworks like TPOT~\citep{Olson2016EvoBio}, AutoGluon~\citep{erickson2020autogluontabularrobustaccurateautoml}, 
and auto-sklearn~\citep{NIPS2015_11d0e628}. By optimizing over a large search space, AutoML frameworks often match the performance of models engineered by machine learning experts~\citep{hutter_automated_2019, erickson2026tabarena}, yet remain constrained to a predefined search space and narrow scope of the end-to-end data science workflow. The recent line of work on tabular foundation models (TFMs) intends to remedy the hyperparameter search and instead train and predict in one forward pass using a pre-trained in-context learning model~\citep{hollmann_accurate_2025, qu_tabicl_2025}. 
Given the time-consuming nature of data science and the limited human capacity, it has naturally become a new objective to automate many steps in the end-to-end data science workflow~\citep{debie2022automatingds}. The increasing code generation capacity of LLMs has led to their deployment in autonomous data science agents, performing tasks across the entire workflow. Approaches for building and refining machine learning pipelines vary significantly. Some agents construct them in a single loop~\citep{grosnit_kolbbased_2025}, while others iteratively refine them~\citep{guo_dsagent_2024a, jiang2025aideaidrivenexplorationspace, zhang_aibuildai_2026} or systematically explore the design space~\citep{du_mlevolve_2026}. Our analysis includes DS-tailored agents, simpler ReAct agents, and AutoGluon as a reference AutoML implementation.

\header{Agentic data science benchmarks and evaluations} Benchmarks have been proposed alongside the proliferation of research into data science agents. Their aim has mostly been to assess the predictive performance of the models these agents produce. These benchmarks predominantly build on machine learning competitions sourced from Kaggle~\citep{chan_mlebench_2025, jing_dsbench_2024, shu_darebench_2025, nie_dsgym_2026, sun2026agenticdatabenchcomprehensivebenchmarkdata, nayak_benchmarking_2026, huang_mlagentbench_2024}, which provide reference solutions that can be used to define relative performance metrics. They focus on predictive performance and mostly do not analyze failures. Those that do report only high-level failure categories, mostly focused on execution errors such as incorrect tool use and coding mistakes~\citep{nie_dsgym_2026, shu_darebench_2025}. We instead analyze \textit{how} data science agents implement predictive models by characterizing which practices they follow and where they fail. 

\header{Failure mode analysis in LLM agents} Analyzing code and execution traces of agentic systems has become essential for understanding these systems. Such analyses have yielded taxonomies of failure modes specific to multi-agent systems~\citep{cemri_why_2026}, safety aspects~\citep{bryan_taxonomy_}, and alignment under adverse incentives~\citep{li2026benchmarkevaluatingoutcomedrivenconstraint}. A consistent finding across analyses is that the ability to write working code and correctly use tools is getting less relevant to the performance of agentic systems, whereas errors concentrate in planning and reasoning~\citep{zhu_why_2026} and understanding data~\citep{zhu_why_2026} and the broader context~\citep{patel2026happensmodeleatsstack}. Existing taxonomies and analyses are mostly proposed for code generation and thus focus on failures in agent coordination, grounding, and execution, providing an incomplete view of violations in data science methodology.

\section{Evaluation Methodology}
\label{sec:method}

In this section, we introduce the methodology used to understand \textit{why} data science agents fail. We first introduce the approach used to carry out large-scale analysis of data science failures using LLM-as-judge (Section~\ref{sec:llmag}), then we present the systematic test suites with curated tests (Section~\ref{sec:uts}), and finally we outline the qualitative comparisons with high-performing solutions (Section~\ref{sec:gap}).

\subsection{Large-scale Failure Mode Analysis \llmjudge}
\label{sec:llmag}

Evaluating agentic data science across a large number of tasks through manual inspection is costly. Following previous evaluations of agentic systems~\citep{cemri_why_2026}, we devise an LLM-as-Judge evaluation~\citep{zheng2023judging} to assess whether generated pipelines exhibit specific failure modes. However, not all failure modes are applicable in every data science use case, and often depend on properties of the task, data, or pipeline itself. For example, ignoring temporal order when splitting the data constitutes a failure \textit{only} when the data involves a temporal structure for which chronological separation is required, and is irrelevant in other cases. 

To account for the task-conditional relevance of failure modes, we apply a two-stage LLM-based evaluation. In the first stage, the judge analyzes each generated solution to identify its pipeline structure and determine which failure modes are applicable. Specifically, it extracts whether the solution contains key pipeline components, including \stage{data cleaning}, \stage{feature engineering}, \stage{validation}, \stage{hyperparameter tuning}, \stage{model comparison}, and \stage{ensembling}, as well as whether it uses deep learning or stochastic models (\stage{uses deep learning} and \stage{uses stochastic model}).
This structural information serves as a filter for determining the applicability of individual failure modes.

In the second stage, the judge evaluates the generated pipeline for the applicable failure modes. We evaluate all failure modes within a single LLM call, providing the pipeline structure extracted in the first stage as additional context. The judge produces a binary decision for each failure mode, indicating whether the corresponding failure is present in the solution, and, if so, extracting the lines of code where the failure mode occurs. Besides filtering out inapplicable failure modes per pipeline, this two-stage evaluation process separates pipeline understanding from failure detection and improves consistent performance across agents. 

We inform the failure modes with commonly identified bad practices~\citep{kumar2025navigatingpitfalls, lones2024avoidingpitfalls}, as shown in Table~\ref{tab:fms_llmag}. These failure modes span the stages of data science pipelines~\cite{}. We broadly categorize them into ``\textit{data leakage}'' modes, indicated using \leakbadge{FM}, and ``\textit{non-leakage}'' modes, using \otherbadge{FM}. Detailed definitions for all failure modes are available in~\Cref {app:failure-modes-definitions}. 
Our analysis focuses on a predefined set of common data science failure modes and therefore does not exhaust all possible issues in a pipeline.  
The detected failures should thus be interpreted as a subset of potential methodological problems.
To partially capture failures outside the taxonomy, we additionally include an \textit{other} category that allows the judge to flag and describe additional issues. We also validate the judge against independent human annotations to assess reliability.

\begin{table}[t]
\vspace{-1.1em}
\centering
\scriptsize
\setlength{\tabcolsep}{3pt}
\caption{Failure modes used for the large-scale analysis. Identifier (ID) colors distinguish two categories: \leakbadge{FM} Data Leakage modes and \otherbadge{FM} Non-leakage modes. Failure modes marked with a dagger ($^\dagger$) are conditionally evaluated based on pipeline stage characteristics (e.g., whether deep learning or ensembling is used).}
\label{tab:fms_llmag}
\begin{tabular}{@{}lp{3.3cm}l@{\hspace{1em}}lp{3.3cm}l@{}}
\toprule
\textbf{ID} & \textbf{Failure} & \textbf{Pipeline Stage} & \textbf{ID} & \textbf{Failure} & \textbf{Pipeline Stage} \\
\midrule
\otherbadge{FM01} & Wrong task formulation & Problem Formulation &
\leakbadge{FM09}  & Feature selection before splitting & Feature Selection \\
\leakbadge{FM02}  & Peeking at test data during EDA & Exploratory Data Analysis &
\otherbadge{FM10} & Wrong model type for task & Model Selection \\
\otherbadge{FM03} & Not cleaning data & Preprocessing &
\otherbadge{FM11} & No regularisation in DL$^\dagger$ & Model Training \\
\leakbadge{FM04}  & Imputation before splitting & Preprocessing &
\otherbadge{FM12} & Naive ensembling$^\dagger$ & Model Optimization \\
\leakbadge{FM05}  & Ignoring temporal order$^\dagger$ & Data Splitting &
\leakbadge{FM13}  & Reusing test set across variants$^\dagger$ & Model Evaluation \\
\otherbadge{FM06}  & No train/val split & Data Splitting &
\otherbadge{FM14} & Single eval of stochastic model$^\dagger$ & Model Evaluation \\
\otherbadge{FM07} & No feature engineering & Feature Engineering &
\otherbadge{FM15} & Using only one eval metric & Model Evaluation \\
\leakbadge{FM08}  & Scaling before splitting & Preprocessing &
& & \\
\bottomrule
\end{tabular}
\vspace{-0.2cm}
\end{table}

\subsection{Curated Test Suite with Data Science Fallacies \testsuite}
\label{sec:uts}

Although the large-scale analysis provides quantitative insights into agent behaviour over naturalistic tasks, it cannot control what the correct practice should look like because the tasks were not designed with specific defects.
To assess agents under more controlled conditions, we design a test suite in which each test targets a specific failure mode through a predefined perturbation. For example, to evaluate whether agents appropriately handle class imbalance, we construct a task with a highly imbalanced target distribution and examine whether the generated pipeline adopts suitable practices. Each test follows a ``unit-test-style" design: the relevant pitfall is deliberately embedded in the task, and failure indicates that the agent did not identify or appropriately address the targeted issue.  For each task, we additionally define multiple levels of hints that vary in instructional explicitness. The base condition provides no hint about the failure mode. We then increase the amount of guidance. 

To determine whether an agent exhibits the appropriate behaviour for a given failure mode, we use an LLM-based judge to inspect the generated pipeline. For each test, we predefine the set of acceptable treatments corresponding to the targeted failure mode, and the judge checks whether any of these treatments is present. To obtain a stable estimation, we run each agent-task-hint combination five times and aggregate the results across runs. The six designed tests are shown in Table~\ref{tab:ts_failures}, with the formalization and evaluation defined in each row. These failure modes cover different stages of data science pipelines. We select these failure modes that satisfy three criteria: (1) can be instantiated through a controlled dataset or task design, (2) the expected methodological response can be specified clearly enough to support a controlled evaluation, and (3) the targeted decision can be isolated from other parts of the pipeline with limited confounding.

\begin{table*}[h]
\centering
\small
\renewcommand{\arraystretch}{1.3}
\caption{Six controlled failure-mode test suites. Each test introduces a targeted perturbation designed to expose a specific data science failure and is evaluated with a strict evaluation measure.}
\label{tab:ts_failures}
\begin{tabularx}{\textwidth}{@{} >{\raggedright\arraybackslash}p{3cm} >{\raggedright\arraybackslash}X >{\raggedright\arraybackslash}X @{}}
\toprule
\textbf{Failure Mode} & \textbf{Test Case} & \textbf{Assessment} \\
\midrule

\tfbadge{TF01} \textbf{Ignoring Class Imbalance}  &  Dataset with high class-imbalance & Assess if the agent is alerting to the skew; classifier performing poorly on minority class  \\
\midrule

\tfbadge{TF02} \textbf{Not Normalizing Formatting}  &
Inconsistent formatting of equivalent values (for example, string-encoded missing values) & 
Assess if an agent is aware of and reacts appropriately to the inconsistency during data preprocessing \\
\midrule

\tfbadge{TF03} \textbf{Not Separating Correlation from Causation}  &
Dataset includes a feature that strongly correlates with the target, without meaningful relationship &  
Assess if an agent is able to distinguish spurious relations by excluding spurious features \\
\midrule

\tfbadge{TF04} \textbf{No Domain Feature Engineering}  &
Prediction task requires a domain-specific feature transformation that is not apparent in the data & 
Assess if an agent is able to apply feature transformation utilizing its pre-trained domain knowledge. \\
\midrule

\tfbadge{TF05} \textbf{Data Leakage in Feature Selection}  &
Dataset includes features
that are not expected to be used at training time.
& 
Assess if an agent is deliberately excluding the features. \\
\midrule

\tfbadge{TF06} \textbf{Wrong Metric Selection}  &
Task with imbalanced classes yielding high accuracy if not evaluated with a proper class-balancing metric & 
Assess if an agent chooses the appropriate metric accounting for imbalance \\

\bottomrule
\end{tabularx}
\end{table*}

\subsection{Skill Gap Analysis with Expert Data Science Solutions \humansolution} 
\label{sec:gap}

The previous evaluations primarily focus on identifying practices that violate established data science principles, hence compromising the validity of a solution. However, the absence of such failures does not necessarily imply that a pipeline is highly effective. 
For example, an agent may produce a methodologically valid pipeline while ignoring a better task formulation, or a non-obvious feature transformation that can better capture the underlying structure of the task.
To understand how data science \textit{agents} compare to \textit{human experts} in data science, we curate a set of Kaggle challenges that accept late submissions. For each task, we submit the predictions or solutions produced by agents to the leaderboard and compare their performance with high-performing solutions. The leaderboard provides an external reference for assessing generated pipelines. 
Moreover, we compare both the predictive performance and the modelling practice used by high-ranking solutions. This helps identify practices that contribute to performance gains even when their absence does not constitute a methodological failure. 
Additionally, this helps in understanding the true performance of agents-generated solutions on realistic challenges.
As an additional reference, we compare the DS agent solutions with the solution of AutoGluon on selected tasks.

\section{Surfacing Silent Failures in Data Science Agents}

Here, we present the insights discovered using the evaluation methods described in~\Cref{sec:method}. We first describe the specific experimental setup used for the evaluation in~\Cref{sec:exp-setup}, then describe the results and main takeaways of the large-scale analysis in~\Cref{sec:large-scale-fm-analysis}, subsequently show the results and insights in~\Cref{sec:res-test-suite}, and finally present the comparisons with expert solutions in~\Cref{sec:res-expert}.

\subsection{Experimental setup}
\label{sec:exp-setup}

\paragraph{Datasets.} 

The evaluation set contains 220 tasks paired with corresponding datasets and is used for the large-scale failure mode analysis \llmjudge. The tasks are drawn from five sources: 108 from STRABLE~\citep{blayer2026strablebenchmarkingtabularmachine}, 44 from Kaggle, 36 from OpenML, 24 from UCI\datasetlinks, and 8 from MLE-bench~\citep{chan_mlebench_2025}. The benchmark comprises 123 regression tasks, of which 122 have a single output and one has multi-outputs, and 97 classification tasks, of which 54 are binary and 43 are multi-class. To prepare the datasets, we perform an 80:20 split for the train and test sets. At runtime, agents are provided with both sets and are expected not to utilize the test set at training time, following the correct practice.
Besides, a separate set of 17 tasks is used for the curated test suite~\testsuite, with 3 underlying datasets. 
Additional dataset details are available in \Cref{app:dataset-details}.

\paragraph{Data science agents.}
We evaluate three data science agents, namely a ReAct baseline~\citep{yao2022react}, AIDE~\citep{jiang2025aideaidrivenexplorationspace}, and MLEvolve~\citep{du_mlevolve_2026}.
MLEvolve is the highest-ranking open-source DS agent in MLEBench~\citep{chan_mlebench_2025}, while AIDE consistently achieved the strongest classification performance in the recent benchmark by \citet{nayak_benchmarking_2026}.
MLEvolve is a self-evolving multi-agent system that iteratively optimizes ML pipelines using methods such as progressive Monte Carlo Graph Search, retrospective memory, and hierarchical adaptive code generation.
AIDE generates pipelines using an iterative code refinement process, leveraging evaluation feedback and execution logs to progressively improve generated solutions over multiple iterations. It consistently achieved the strongest classification performance in the recent benchmark by \citet{nayak_benchmarking_2026}.
For comparison, we set both AIDE and MLEvolve agents' steps to 10.
Additionally, we include AutoGluon~\citep{erickson2020autogluontabularrobustaccurateautoml}, a widely used AutoML system, which serves as a strong non-agentic baseline based on heuristic-driven optimization and ensemble learning. Comparing these solutions across different failure modes allows us to obtain insights into what agent frameworks are really improving. 

\paragraph{LLMs.}
The capacity of LLMs is important to agentic systems, since smaller models usually perform worse than larger LLMs in agentic tasks.  To understand the impact of different models within agentic data science frameworks, we use three models of different capacities. We select \qwenSmall and \qwenMedium, which can be hosted locally on commodity hardware, and \texttt{gemini-3.7-flash}, which we select because it has a good performance-efficiency trade-off in agentic tasks ~\cite{google2026gemini37flash}. 
%
All LLM judges use the more capable \texttt{gemini-3.1-pro}. Additionally, for the curated test suite, we also evaluate \gptSol.
To assess the reliability of the LLM judge output, we compare with independent human annotations. Across 285 failure mode ratings among 19 pipelines, the judge achieves 89.5\% agreement with the human annotator (Cohen's $\kappa=0.62$, 95\% CI [0.50, 0.73]), with precision (0.71) and recall (0.65) for detecting present failures. The detailed agreement breakdown is in Appendix~\ref{app:human}.

\header{Validation of dataset memorization robustness} Many tabular prediction tasks are publicly available online, and LLMs are likely to have seen them in their training data. To understand if a good predictive performance is partly because the underlying model has seen the tabular data, we run memorization tests using column header completion~\citep{bordt2024elephants, lai2026kramabench}. We then select a subset of tasks with high and low memorization rates, respectively. Following a similar procedure as in KramaBench~\citep{lai2026kramabench}, we then randomize the column names and rerun the agents on tasks with masked metadata and description. We also rerun the agents on the unmasked version as a placebo to obtain unbiased effects. A large difference in $\Delta_{plc}$ and $\Delta_{alt}$ for a high-memorization set indicates that the performance is derived from memorization and vice versa. 
Finally, we apply the non-parametric Wilcoxon test to compare whether there are significant predictive performance differences when tables are memorized. We test this on three DS agent frameworks paired with \geminiFlash{3.7} since it has the lowest data science failure rate. 
We present the results in~\Cref{app:memorization-experiment}. All $p$ values are non-significant in both sets with high and low memorization rates, indicating that memorizing tables does not significantly affect predictive performance.

\vspace{-.5em} 

\subsection{Large-scale Analysis Surfaces Recurrent Silent Failures \llmjudge}
\label{sec:large-scale-fm-analysis}
\vspace{-.5em}

Table~\ref{tab:judge-by-agent-fm-x-heatmap} shows the prevalence of failure modes in agent frameworks and models. It shows that the specific failures the pipelines exhibit depend on how completely they cover the data science workflow. Failures happen either because agents perform operations incorrectly, or because they outright skip parts of the workflow (e.g., ``\textit{No val split}'', ``\textit{No feature engineering}''). More capable models and elaborate frameworks shift failures from skipping steps to failures within these steps.

\begin{table*}
\centering
\scriptsize
\setlength{\tabcolsep}{0.5pt}
\begin{tabular}{llccccccccccccccc}
\toprule
\textbf{Agent} & \textbf{Model} & 
\rotatebox{75}{\tiny \textcolor{bluetext}{Wrong task formulation}} & 
\rotatebox{75}{\tiny \textcolor{redtext}{Peeking at test data during EDA}} & 
\rotatebox{75}{\tiny \textcolor{bluetext}{Not cleaning data}} & 
\rotatebox{75}{\tiny \textcolor{redtext}{Imputation before splitting}} & 
\rotatebox{75}{\tiny \textcolor{redtext}{Ignoring temporal order}} & 
\rotatebox{75}{\tiny \textcolor{bluetext}{No train/val split}} & 
\rotatebox{75}{\tiny \textcolor{bluetext}{No feature engineering}} & 
\rotatebox{75}{\tiny \textcolor{redtext}{Scaling before splitting}} & 
\rotatebox{75}{\tiny \textcolor{redtext}{Feature selection before splitting}} & 
\rotatebox{75}{\tiny \textcolor{bluetext}{Wrong model type for task}} & 
\rotatebox{75}{\tiny \textcolor{bluetext}{No regularisation in DL}} & 
\rotatebox{75}{\tiny \textcolor{bluetext}{Naive ensembling}} & 
\rotatebox{75}{\tiny \textcolor{redtext}{Reusing test set across variants}} & 
\rotatebox{75}{\tiny \textcolor{bluetext}{Single eval of stochastic model}} & 
\rotatebox{75}{\tiny \textcolor{bluetext}{Using only one eval metric}} \\
\midrule
\multirow{3}{*}{\textbf{ReAct}}  & Qwen-4B & \cellcolor[HTML]{FEE7C5}{\color[HTML]{000000}0.02} & \cellcolor[HTML]{FEECD1}{\color[HTML]{000000}0.01} & \cellcolor[HTML]{F98455}{\color[HTML]{000000}0.28} & \cellcolor[HTML]{FEEFD9}{\color[HTML]{000000}0.00} & \cellcolor[HTML]{FFFFFF}{\color[HTML]{000000}0.00} & \cellcolor[HTML]{7F0000}{\color[HTML]{FFFFFF}1.00} & \cellcolor[HTML]{820000}{\color[HTML]{FFFFFF}0.98} & \cellcolor[HTML]{FFFFFF}{\color[HTML]{000000}0.00} & \cellcolor[HTML]{FFFFFF}{\color[HTML]{000000}0.00} & \cellcolor[HTML]{FEE9CB}{\color[HTML]{000000}0.01} & \cellcolor[HTML]{D9D9D9}{\color[HTML]{666666}N/A} & \cellcolor[HTML]{D9D9D9}{\color[HTML]{666666}N/A} & \cellcolor[HTML]{D9D9D9}{\color[HTML]{666666}N/A} & \cellcolor[HTML]{E14630}{\color[HTML]{FFFFFF}0.48} & \cellcolor[HTML]{FDB982}{\color[HTML]{000000}0.15} \\
 & Qwen-27B & \cellcolor[HTML]{FEE9CB}{\color[HTML]{000000}0.01} & \cellcolor[HTML]{FDC089}{\color[HTML]{000000}0.12} & \cellcolor[HTML]{FDDCAF}{\color[HTML]{000000}0.04} & \cellcolor[HTML]{FDBE87}{\color[HTML]{000000}0.13} & \cellcolor[HTML]{F26E4C}{\color[HTML]{000000}0.36} & \cellcolor[HTML]{FDDAAB}{\color[HTML]{000000}0.05} & \cellcolor[HTML]{DF412C}{\color[HTML]{FFFFFF}0.50} & \cellcolor[HTML]{FFFFFF}{\color[HTML]{000000}0.00} & \cellcolor[HTML]{FEE0B7}{\color[HTML]{000000}0.03} & \cellcolor[HTML]{FEEFD9}{\color[HTML]{000000}0.00} & \cellcolor[HTML]{D9D9D9}{\color[HTML]{666666}N/A} & \cellcolor[HTML]{E0442F}{\color[HTML]{FFFFFF}0.49} & \cellcolor[HTML]{FFFFFF}{\color[HTML]{000000}0.00} & \cellcolor[HTML]{890000}{\color[HTML]{FFFFFF}0.95} & \cellcolor[HTML]{960000}{\color[HTML]{FFFFFF}0.89} \\
 & Gemini-3.7 & \cellcolor[HTML]{FEE4C0}{\color[HTML]{000000}0.02} & \cellcolor[HTML]{FDD29C}{\color[HTML]{000000}0.07} & \cellcolor[HTML]{F77D52}{\color[HTML]{000000}0.30} & \cellcolor[HTML]{FDD5A0}{\color[HTML]{000000}0.06} & \cellcolor[HTML]{E24831}{\color[HTML]{FFFFFF}0.48} & \cellcolor[HTML]{EE6246}{\color[HTML]{FFFFFF}0.40} & \cellcolor[HTML]{F06849}{\color[HTML]{000000}0.38} & \cellcolor[HTML]{FEEFD9}{\color[HTML]{000000}0.00} & \cellcolor[HTML]{FDCD96}{\color[HTML]{000000}0.08} & \cellcolor[HTML]{FEEFD9}{\color[HTML]{000000}0.00} & \cellcolor[HTML]{D9D9D9}{\color[HTML]{666666}N/A} & \cellcolor[HTML]{930000}{\color[HTML]{FFFFFF}0.90} & \cellcolor[HTML]{D9D9D9}{\color[HTML]{666666}N/A} & \cellcolor[HTML]{FDDBAD}{\color[HTML]{000000}0.04} & \cellcolor[HTML]{FDBC85}{\color[HTML]{000000}0.14} \\
\midrule
\multirow{3}{*}{\textbf{AIDE}}  & Qwen-4B & \cellcolor[HTML]{FEE9CA}{\color[HTML]{000000}0.01} & \cellcolor[HTML]{FEE6C4}{\color[HTML]{000000}0.02} & \cellcolor[HTML]{BB0B07}{\color[HTML]{FFFFFF}0.72} & \cellcolor[HTML]{FDCF99}{\color[HTML]{000000}0.08} & \cellcolor[HTML]{FDBF88}{\color[HTML]{000000}0.13} & \cellcolor[HTML]{FDD19B}{\color[HTML]{000000}0.07} & \cellcolor[HTML]{8C0000}{\color[HTML]{FFFFFF}0.93} & \cellcolor[HTML]{FEE6C4}{\color[HTML]{000000}0.02} & \cellcolor[HTML]{FFFFFF}{\color[HTML]{000000}0.00} & \cellcolor[HTML]{FEDFB5}{\color[HTML]{000000}0.03} & \cellcolor[HTML]{D9D9D9}{\color[HTML]{666666}N/A} & \cellcolor[HTML]{D9D9D9}{\color[HTML]{666666}N/A} & \cellcolor[HTML]{D9D9D9}{\color[HTML]{666666}N/A} & \cellcolor[HTML]{FDA56F}{\color[HTML]{000000}0.19} & \cellcolor[HTML]{9B0000}{\color[HTML]{FFFFFF}0.87} \\
 & Qwen-27B & \cellcolor[HTML]{FEE2BB}{\color[HTML]{000000}0.03} & \cellcolor[HTML]{FDCF99}{\color[HTML]{000000}0.08} & \cellcolor[HTML]{FC9C67}{\color[HTML]{000000}0.21} & \cellcolor[HTML]{FEE2BB}{\color[HTML]{000000}0.03} & \cellcolor[HTML]{FDB67F}{\color[HTML]{000000}0.15} & \cellcolor[HTML]{FEEFD8}{\color[HTML]{000000}0.01} & \cellcolor[HTML]{BB0B07}{\color[HTML]{FFFFFF}0.71} & \cellcolor[HTML]{FFFFFF}{\color[HTML]{000000}0.00} & \cellcolor[HTML]{FEE8C8}{\color[HTML]{000000}0.02} & \cellcolor[HTML]{FFFFFF}{\color[HTML]{000000}0.00} & \cellcolor[HTML]{D9D9D9}{\color[HTML]{666666}N/A} & \cellcolor[HTML]{A60000}{\color[HTML]{FFFFFF}0.81} & \cellcolor[HTML]{D9D9D9}{\color[HTML]{666666}N/A} & \cellcolor[HTML]{FDCC96}{\color[HTML]{000000}0.08} & \cellcolor[HTML]{9E0000}{\color[HTML]{FFFFFF}0.85} \\
 & Gemini-3.7 & \cellcolor[HTML]{FEECD1}{\color[HTML]{000000}0.01} & \cellcolor[HTML]{FDB67F}{\color[HTML]{000000}0.15} & \cellcolor[HTML]{E55038}{\color[HTML]{FFFFFF}0.45} & \cellcolor[HTML]{FDBF88}{\color[HTML]{000000}0.13} & \cellcolor[HTML]{E44D35}{\color[HTML]{FFFFFF}0.47} & \cellcolor[HTML]{FFFFFF}{\color[HTML]{000000}0.00} & \cellcolor[HTML]{FDC18A}{\color[HTML]{000000}0.12} & \cellcolor[HTML]{FEE0B7}{\color[HTML]{000000}0.03} & \cellcolor[HTML]{FDBA83}{\color[HTML]{000000}0.14} & \cellcolor[HTML]{FFFFFF}{\color[HTML]{000000}0.00} & \cellcolor[HTML]{D9D9D9}{\color[HTML]{666666}N/A} & \cellcolor[HTML]{E54E36}{\color[HTML]{FFFFFF}0.46} & \cellcolor[HTML]{FFFFFF}{\color[HTML]{000000}0.00} & \cellcolor[HTML]{FEDFB5}{\color[HTML]{000000}0.03} & \cellcolor[HTML]{990000}{\color[HTML]{FFFFFF}0.88} \\
\midrule
\multirow{3}{*}{\textbf{MLEvolve}}  & Qwen-4B & \cellcolor[HTML]{FEE8C9}{\color[HTML]{000000}0.02} & \cellcolor[HTML]{FEE8C9}{\color[HTML]{000000}0.02} & \cellcolor[HTML]{9C0000}{\color[HTML]{FFFFFF}0.86} & \cellcolor[HTML]{FDCA93}{\color[HTML]{000000}0.09} & \cellcolor[HTML]{FEE4C0}{\color[HTML]{000000}0.02} & \cellcolor[HTML]{BE0F0A}{\color[HTML]{FFFFFF}0.69} & \cellcolor[HTML]{8C0000}{\color[HTML]{FFFFFF}0.94} & \cellcolor[HTML]{FEE8C9}{\color[HTML]{000000}0.02} & \cellcolor[HTML]{FFFFFF}{\color[HTML]{000000}0.00} & \cellcolor[HTML]{FFFFFF}{\color[HTML]{000000}0.00} & \cellcolor[HTML]{D9D9D9}{\color[HTML]{666666}N/A} & \cellcolor[HTML]{D9D9D9}{\color[HTML]{666666}N/A} & \cellcolor[HTML]{D9D9D9}{\color[HTML]{666666}N/A} & \cellcolor[HTML]{7F0000}{\color[HTML]{FFFFFF}1.00} & \cellcolor[HTML]{F4734E}{\color[HTML]{000000}0.34} \\
 & Qwen-27B & \cellcolor[HTML]{FEE8C9}{\color[HTML]{000000}0.01} & \cellcolor[HTML]{FDCE98}{\color[HTML]{000000}0.08} & \cellcolor[HTML]{FDAF79}{\color[HTML]{000000}0.17} & \cellcolor[HTML]{FEE0B8}{\color[HTML]{000000}0.03} & \cellcolor[HTML]{FDAD77}{\color[HTML]{000000}0.17} & \cellcolor[HTML]{FEEBD0}{\color[HTML]{000000}0.01} & \cellcolor[HTML]{FDB37D}{\color[HTML]{000000}0.16} & \cellcolor[HTML]{FEEFD9}{\color[HTML]{000000}0.00} & \cellcolor[HTML]{FEEBD0}{\color[HTML]{000000}0.01} & \cellcolor[HTML]{FEEFD9}{\color[HTML]{000000}0.00} & \cellcolor[HTML]{FFFFFF}{\color[HTML]{000000}0.00} & \cellcolor[HTML]{DE3F2B}{\color[HTML]{FFFFFF}0.51} & \cellcolor[HTML]{FDD5A0}{\color[HTML]{000000}0.06} & \cellcolor[HTML]{B30000}{\color[HTML]{FFFFFF}0.76} & \cellcolor[HTML]{990000}{\color[HTML]{FFFFFF}0.87} \\
 & Gemini-3.7 & \cellcolor[HTML]{FEEFD9}{\color[HTML]{000000}0.00} & \cellcolor[HTML]{FDC690}{\color[HTML]{000000}0.10} & \cellcolor[HTML]{FDC690}{\color[HTML]{000000}0.10} & \cellcolor[HTML]{FDC892}{\color[HTML]{000000}0.10} & \cellcolor[HTML]{F26E4C}{\color[HTML]{000000}0.36} & \cellcolor[HTML]{FEEFD9}{\color[HTML]{000000}0.00} & \cellcolor[HTML]{FEEFD9}{\color[HTML]{000000}0.00} & \cellcolor[HTML]{FEE9CB}{\color[HTML]{000000}0.01} & \cellcolor[HTML]{FDD5A0}{\color[HTML]{000000}0.06} & \cellcolor[HTML]{FFFFFF}{\color[HTML]{000000}0.00} & \cellcolor[HTML]{FFFFFF}{\color[HTML]{000000}0.00} & \cellcolor[HTML]{FDDBAC}{\color[HTML]{000000}0.04} & \cellcolor[HTML]{FEEDD4}{\color[HTML]{000000}0.01} & \cellcolor[HTML]{FDBD86}{\color[HTML]{000000}0.13} & \cellcolor[HTML]{E9583E}{\color[HTML]{FFFFFF}0.43} \\
\bottomrule
\end{tabular}
\caption{Occurrence rate per failure modes (FM) by agent and model over the pipelines for which the FM is applicable. N/A indicates that the number of applicable pipelines (as defined in Section~\ref{sec:large-scale-fm-analysis}) for a given FM is less than 20. Red indicates leakage modes and blue indicates non-leakage modes.}
\label{tab:judge-by-agent-fm-x-heatmap}
\vspace{-0.4cm}
\end{table*}

\header{Failures occur across models and frameworks} All framework--model configurations have at least one failure mode that occurs in 43\% or more of the pipelines. Agent frameworks reduce some of the failures of the ReAct baseline. With \geminiFlash{3.7}, both AIDE and MLEvolve reduce ``\textit{No val split}'' from 0.40 to 0.00, and MLEvolve reduces ``\textit{No feature engineering}'' from 0.38 to 0.00 and ``\textit{naive ensembling}'' from 0.90 to 0.04. At the same time, frameworks also increase non-leakage failure modes. AIDE reports only a single evaluation metric in 85\%--88\% of its pipelines regardless of the model, compared with 15\% and 14\% for ReAct with \qwenSmall and \geminiFlash{3.7}, which is consistent with its design of scoring and selecting candidate solutions by a single validation metric~\citep{jiang2025aideaidrivenexplorationspace}. Whether a framework improves adherence to DS practices depends on the model used. For example, AIDE lowers ``\textit{No feature engineering}'' with \geminiFlash{3.7} (0.38 to 0.12), but raises it with \qwenMedium (0.50 to 0.71).

\header{Higher capacity models cover the data-science stages better but do not follow good practices}
The low rates of leakage failures for small models do not indicate sound practices. ReAct with \qwenSmall never designates a validation set and rarely performs feature engineering; consequently, the leakage modes that depend on the agent performing either do not occur (``\textit{Impute before split}'', ``\textit{Ignores temporal order}'', ``\textit{Scale before split}'', and ``\textit{Feature-selection before split}'' are all 0.00). Under AIDE, \qwenSmall does use a validation split, resulting in increased leakage across the related failure mode; for instance, \textit{imputation before splitting} increases from 0.00 to 0.08. More capable models more frequently perform feature engineering and designate a validation set themselves, but they similarly show elevated leakage failure modes. For example, \geminiFlash{3.7} ignores the temporal order in more than a third of the pipelines over applicable tasks for all three agents. Overall, the results indicate that model capacity improves the \textit{coverage} of the data-science workflow but that higher capacity models still commit errors within the expanded workflow. Thus, the raw number of failures over a pipeline is not indicative of the pipelines quality, since skipping part of the workflow removes the opportunity to fail within it.

\begin{wrapfigure}{r}{0.43\textwidth} 
    \vspace{-1.0em}
    \centering
    \includegraphics[width=\linewidth]{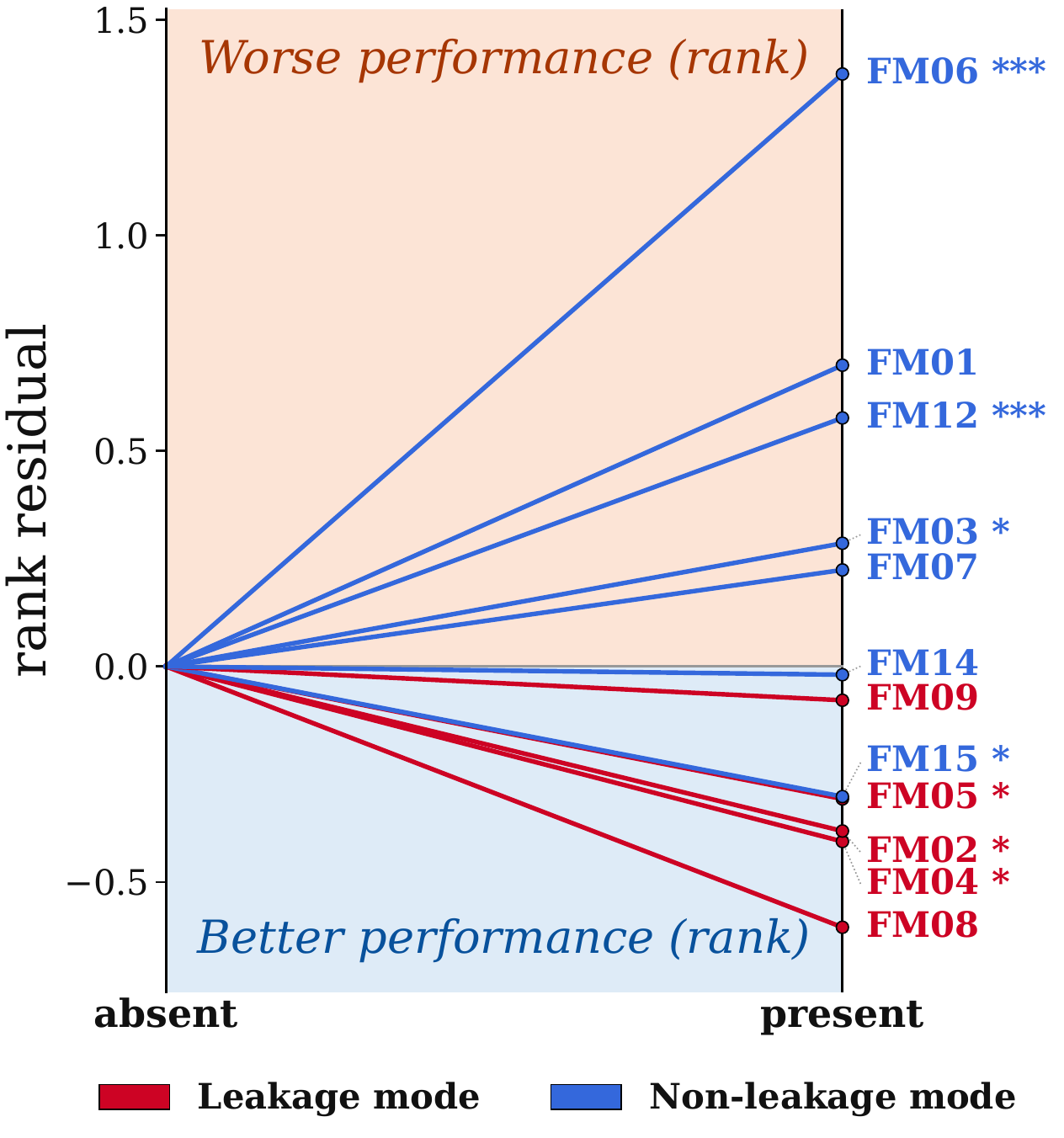}
    \vspace{-0.4cm}
    \caption{
        Adjusted effect of FMs on residual rank in the \textit{test} set that is accessible to the agent. A high rank indicates worse performance, while a low rank indicates better performance (* indicates significance). 
    }
    \label{fig:perf_judge}
    \vspace{-1em}
\end{wrapfigure}

\header{Leakage extends beyond the taxonomy}
We additionally analyse the failures flagged by the judge that fall outside of the fixed taxonomy, with full details reported in~\Cref{app:fm_beyond_taxonomy}.
We find additional failures in 37\% of the pipelines. 
Most commonly in the form of further leakage modes, namely leakage of validation into train data, e.g., where preprocessing or target encoding is fitted before cross-validation, and additional leakage of test into train data.
The leakage modes show increasing prevalence with increasing model capacity, following the same trend as the leakage modes inside the taxonomy. 
Notably, 50\% of the pipelines of MLEvolve with \geminiFlash{3.7}, the configuration with the fewest failures within the taxonomy, exhibit leakage between validation and train data, even though MLEvolve includes a dedicated agent for checking leakages~\mbox{\citep{du_mlevolve_2026}}.

\header{Failure types correlate with reported performance}
Lastly, we analyse how the failures relate to predictive performance. For each task, we rank the pipelines of each framework--model configuration by the predictive performance the pipelines achieve on the test set \textit{provided to the agent}.
Since the configuration (i.e., a combination of model and the agent framework) influences the failures that occur in a pipeline as well as its performance, a direct comparison of raw ranks would primarily reflect differences between these configurations. We therefore compare each pipeline's rank to the rank expected for its configuration, and measure how this difference changes when a failure mode is present (\Cref{sec:residual-rank-details}). 
Figure~\ref{fig:perf_judge} shows that leakage and non-leakage modes affect performance in opposite directions. \textit{All data leakage modes lead to improved rankings, which can be expected as leakage introduces information from the evaluation data into the training process.} This illustrates why downstream performance alone is not a good measure of agentic DS capability. The non-leakage failure modes have the opposite relation: more failure modes correlate with lower ranks. Not using a validation split (\otherbadge{FM06}) to assess model generalizability has the largest effect, deteriorating the performance by more than a full rank position on average. An exception to the pattern is that using only one evaluation metric (\otherbadge{FM15}) is associated with a better rank. We hypothesize that pipelines that evaluate a single metric optimize narrowly for the metric specified by the task, which is also the metric we rank by, possibly at the cost of robustness that a single metric cannot capture. Overall, the results show that a better score on test data does not imply a better pipeline, and evaluating agents solely by their scores can reward leakage modes.

\subsection{Controlled Evaluation Confirms Consistent Failure Patterns \testsuite} 
\label{sec:res-test-suite}

\begin{figure}[b]
    \centering
    \vspace{-0.5cm}
    \begin{minipage}[t]{0.69\textwidth}
        \centering
        \includegraphics[width=\linewidth]{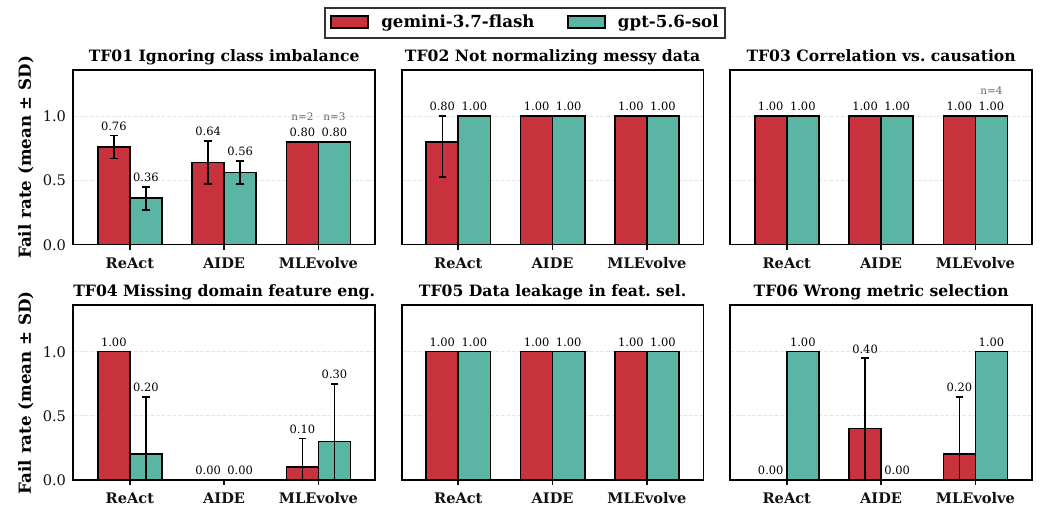}
        \caption{Results for the curated test suite of failure modes. 
        Failure rate represents failure occurrences per failure mode, averaged over $n=5$ runs. A failure rate of 1.0 means it occurs in every run.}
        \label{fig:dsts_fm}
    \end{minipage}\hfill
    \begin{minipage}[t]{0.29\textwidth}
        \centering
        \includegraphics[width=\linewidth]{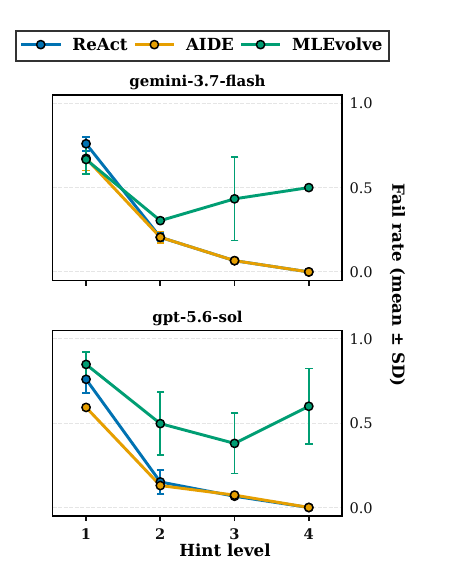}
        \caption{Average FMs per hint level. Hint level 1 is no hint; higher is more explicit.
        }
        \label{fig:dsts_hint}
    \end{minipage}
\end{figure}

Figure~\ref{fig:dsts_fm} reports failure rates on the six controlled tests for ReAct, AIDE, and MLEvolve using \geminiFlash{3.7} and \gptSol. Figure~\ref{fig:dsts_hint} further examines how these failures change as increasingly explicit hints are provided. We also evaluated \qwenSmall and \qwenMedium; however, most runs failed to produce usable executable pipelines and therefore are excluded from the analysis. Next, we discuss our main findings in turn.

\header{Some methodological failures persist across configurations} The clearest examples are correlation and causation reasoning (\tfbadge{TF03}) and leakage in feature selection (\tfbadge{TF05}), where every configuration reaches a failure rate of $1.0$. Messy data normalization (\tfbadge{TF02}) shows similar behaviour: five configurations fail every run, while ReAct with \geminiFlash{3.7} fails at a rate of $0.8$. In contrast, class imbalance (\tfbadge{TF01}), domain feature engineering (\tfbadge{TF04}), and metric selection (\tfbadge{TF06}) vary substantially across agents and models, indicating that different methodological decisions depend on both the underlying LLM and the agent framework.

\header{Stronger models do not consistently yield better data science practice} Compared to \geminiFlash{3.7}, \gptSol reduces failures for ReAct on class imbalance (\tfbadge{TF01}) and domain feature engineering (\tfbadge{TF04}), but the effect does not generalize across tests. This suggests that improvements in model capability do not transfer uniformly to methodological decision-making.

\header{Explicit hints reduce failures, but not all} Figure~\ref{fig:dsts_fm} shows that ReAct and AIDE improve sharply as hints become more explicit, with failure rates approaching zero under the strongest guidance. This suggests that many failures arise from failing to recognize the relevant methodological constraint rather than from an inability to carry out the correct practice once prompted. However, MLEvolve behaves differently: although moderate hints reduce its failure rate, stronger hints do not lead to further improvement, and its failure rate remains substantially higher than ReAct and AIDE under more explicit guidance. 
\textit{This points to a limitation of more advanced frameworks, where elaborate search and optimization may fail to maintain explicit constraints during the pipeline generation.}

\subsection{Data Science Agents Rarely Discover Expert-level Strategies \humansolution} 
\label{sec:res-expert}

Table~\ref{tab:hsc-results} compares agent-generated pipelines with AutoGluon solutions and the 1st-place Kaggle solutions across five challenges. Expert solutions remain ahead, and the performance gap varies by challenge. On the two Playground competitions, Heart Disease Prediction and Obesity risk, the gap is relatively small, while larger gaps remain on Single-Cell Perturbation, Problematic Internet Use, and Survival Predictions. This suggests that automated pipelines can be competitive on relatively accessible tabular tasks, while more specialized challenges leave large room for improvements.

\begin{table*}[h]
\centering
\scriptsize
\setlength{\tabcolsep}{4pt}
\renewcommand{\arraystretch}{1.15}
\caption{Leaderboard performance across Kaggle challenges. Scores reported as \textbf{private / public}. Best overall score is bolded, and the second-best is underscored. AutoGluon solutions are applicable only for challenges with CSV uploads (i.e., heart disease prediction and obesity risk) which are playground tasks and hence more simplistic in nature.
}
\begin{tabular}{ll ccccc}
\toprule
& & \multicolumn{3}{c}{\textbf{Data Science Agents}} & \textbf{AutoML} & \textbf{Baseline} \\
\cmidrule(lr){3-5} \cmidrule(lr){6-6} \cmidrule(lr){7-7}
\textbf{Challenge} & \textbf{Metric} & \textbf{ReAct} & \textbf{AIDE} & \textbf{MLEvolve} & \textbf{AutoGluon} & \textbf{1st Place (Human)} \\
\midrule
Single-Cell Perturbations & MRRMSE $\downarrow$ 
    & 1.1020 / 0.7690 
    & 0.8970 / 0.6600 
    & \underline{0.8310} / \underline{0.6170} 
    & -- / -- 
    & \textbf{0.7290} / \textbf{0.5230} \\

Heart Disease Prediction & AUC-ROC $\uparrow$  
    & 0.9547 / 0.9531 
    & 0.9549 / 0.9533 
    & 0.9546 / 0.9531 
    & \underline{0.9551} / \underline{0.9536} 
    & \textbf{0.9554} / \textbf{0.9542} \\

Obesity Risk & Accuracy $\uparrow$ 
    & 0.9065 / \underline{0.9090} 
    & 0.9059 / 0.9025 
    & 0.9071 / 0.9035 
    & \underline{0.9074} / 0.9075 
    & \textbf{0.9116} / \textbf{0.9234} \\

Problematic Internet Use & QWK $\uparrow$ 
    & \underline{0.4440} / \underline{0.4680} 
    & 0.4280 / 0.4640 
    & 0.4170 / 0.4480 
    & -- / -- 
    & \textbf{0.4820} / \textbf{0.5010} \\

Survival Predictions & C-index $\uparrow$ 
    & 0.6502 / 0.6465 
    & \underline{0.6536} / \underline{0.6488} 
    & 0.3155 / 0.3233 
    & -- / -- 
    & \textbf{0.7013} / \textbf{0.6981} \\
\bottomrule
\end{tabular}
\label{tab:hsc-results}
\end{table*}

\header{No agent dominates across tasks} MLEvolve performs best among the agents on Single-Cell Perturbations, ReAct on Problematic Internet Use, and AIDE on Survival Predictions. These results suggest that improvements introduced by the frameworks are not universal. Interestingly, MLEvolve performs poorly on Survival Predictions because it optimized on the inverted ranking, illustrating how a single error in decision-making can dominate downstream performance. 

\header{Case Study}
To examine where this gap comes from, we compare the pipelines generated by ReAct, AIDE, and MLEvolve using \geminiFlash{3.7} with the 1st-place solution for the Kaggle \textit{Problematic Internet Use} challenge (Figure~\ref{fig:combined-dags} in Appendix~\ref{app:case-study}). Our test split contains \texttt{PCIAT-*} variables, including \texttt{PCIAT-PCIAT\_Total}, from which the target \texttt{sii} can be derived, although these variables are not included in the actual Kaggle test set. ReAct and AIDE selected these leaked features and consequently achieved local QWK scores near 1.0, despite producing pipelines that would fail under the competition setting. \textit{This again confirms that predictive scores during agents' optimization are not trustworthy.} Structural evaluation further reveals recurring weaknesses across frameworks: ReAct exhibited preprocessing leakage, limited feature engineering, arbitrary ensemble weights, and invalid test IDs; AIDE relied largely on raw features and mishandled categorical variables; and MLEvolve produced richer domain-informed features but still introduced leakage during encoding and relied on equal-weight ensembling. But this case study also indicates that \textit{the gap to expert solutions goes beyond failures.} The 1st-place solution has fundamentally different, creative modelling choices. For example, it predicts the continuous \texttt{PCIAT\_Total} rather than \texttt{sii} directly, applied optimized thresholds only at the final stage, used a Tweedie objective, adopted multi-seed 10-fold cross-validation, and combined diverse model families. These findings can inform future DS agent design to also focus on higher-level choices in problem formulation, task-specific objectives, validation design, feature engineering, and model diversity.

\section{Conclusion}
With the strong coding capabilities of large language models, they are increasingly integrated in data science agents. With higher-capacity models and advanced data science scaffolding, recent benchmarks show improving downstream performance. In this work, \textit{we systematically evaluate failure modes in the pipelines that DS agents implement to get a more nuanced view of their true data science practices beyond accuracy}. We perform a large-scale analysis across 220 tasks and nine configurations of model--agent configurations, an extensible controlled test suite, and comparisons with expert real-world solutions. We surface many different failures that purely downstream performance evaluation masks. In fact, our results indicate that data leakage failure modes -- which we identify to exist in many forms -- can actually lead to stronger predictive performance on a test set.
We also find that more sophisticated agent frameworks do not necessarily eliminate these failures. Despite the presence of dedicated data leakage checks in some frameworks, these can still exhibit leakage failures.
This suggests that preventing such failures requires more than adding specialized components. Agent designs must reason more explicitly about methodological validity throughout the pipeline. Altogether, our findings reinforce the need to evaluate agentic data science \textit{beyond accuracy}.
We envision that our findings will improve agentic DS evaluations broadly, and inform the design of future agentic data science systems and models.

\section*{AI use statement}

In this work, we used generative AI tools to implement methods, to clean and reformat datasets, to support qualitative and thematic data analysis, and to interpret results.
We have not used generative AI tools to design or provide feedback on research methodology or experiments, to generate synthetic data sets, to help develop theoretical models or conceptual frameworks, to propose or refine hypotheses,
and to assist in the writing of proofs, to formulate mathematical claims, to provide critical ingredients for proving mathematical claims, and to assist with translation are not applicable to this work. 
Additionally, we used generative AI tools to create or modify scientific figures, suggest experimental parameters, create and edit software code, create artifacts such as traces and LLM-judge logs evaluated in this project, and to edit the research paper to improve readability. 
We have reviewed all AI-assisted work. We validated each of the generated artifacts, and thoroughly reviewed and tested the generated code by one author. We validated every generated figure and assessed its validity. We take responsibility for the final content of this work, including text, claims or artifacts produced with the aid of generative AI.

\section*{Acknowledgements}
This work was supported by the Dutch Research Council (NWO, AiNed Fellowship grant NGF.1607.22.045) and a grant from SAP. We also gratefully acknowledge support through compute credit grants from Google and OpenAI. This work also used the Dutch national e-infrastructure with the support of the SURF Cooperative using grant no. EINF-19960.

\bibliography{iclr2027_conference}
\bibliographystyle{iclr2027_conference}

\newpage
\appendix
\crefalias{section}{appendix}
\crefalias{subsection}{appendix}
\section{Appendix}

\subsection{Details on memorization check setup}
\label{app:memorization-experiment}

\Cref{tab:mem-test} presents the $p$ values for the paired Wilcoxon test on both the high-memorised set and low-memorised set. The testing shows that the performance has no significant changes regardless of memorizing the tables. 

\begin{table}[h]
\centering
\caption{Wilcoxon signed-rank test of $\Delta_{alt}$ vs.\ the rerun randomized $\Delta_{plc}$, computed separately for high- and low-memorization rate tasks across agentic frameworks. None of the six tests reaches significance at $\alpha = 0.05$, so dataset memorization does not measurably influence performance.}
\label{tab:mem-test}
\begin{tabular}{lrrrr}
\toprule
Agent & $n$ (high) & Wilcoxon $p$ (high) & $n$ (low) & Wilcoxon $p$ (low) \\
\midrule
ReAct  & 17 & 0.890 & 18 & 0.212 \\
AIDE     & 16 & 0.231 & 16 & 0.782 \\
MLEvolve & 16 & 0.597 & 17 & 0.890 \\
\bottomrule
\end{tabular}
\end{table}

\subsection{Failure Modes Definitions}
\label{app:failure-modes-definitions}
In \Cref{tab:fm-taxonomy} we include the definition of each of the failure modes evaluated in the large-scale analysis.

\begin{table}[h]
\centering
\small
\begin{tabular}{@{}llp{8.0cm}@{}}
\toprule
\textbf{ID} & \textbf{Failure Mode} & \textbf{Definition} \\
\midrule
FM01 & Wrong task formulation & The agent frames the modelling problem in a way that discards the signal the task actually provides. \\
FM02 & Peeking at test data during EDA & Exploring test data during initial analysis creates untestable assumptions baked into model design. \\
FM03 & Not cleaning data & Ignoring data duplication, missing values, and outliers/inconsistencies before model building. \\
FM04 & Imputation before splitting & Performing mean/median imputation before splitting -- leaks test statistics into training. \\
FM05 & Ignoring temporal order & Randomly splitting time-series data without respecting temporal order -- introduces look-ahead bias. \\
FM06 & No train/val split & Model trained and evaluated on the same data -- performance is a misleading overestimate. \\
FM07 & No feature engineering & The agent uses raw columns as-is (at most basic imputation/encoding) with no derived features or domain-informed transforms, even where the task or column names suggest obvious ones. \\
FM08 & Scaling before splitting & Fitting scaler on the full dataset before splitting -- test statistics leak into training. \\
FM09 & Feature selection before splitting & Selecting features on the full dataset before splitting -- causes test leakage. \\
FM10 & Wrong model type for task & Using a regression model for a classification task (or vice versa) -- conceptual mismatch. \\
FM11 & No regularisation in DL & Training deep learning models without dropout or weight decay -- prone to overfitting. \\
FM12 & Naive ensembling & When the agent combines multiple models, it uses an unweighted/equal-weight average or simple majority vote with no tuning of the combination itself, where a more sophisticated combination is straightforward given what's already computed. \\
FM13 & Reusing test set across variants & Same test set used to select among multiple models -- test set gradually leaks into model selection (sequential overfitting). \\
FM14 & Single eval of stochastic model & Single evaluation of a stochastic/unstable model; results may not reflect true performance. \\
FM15 & Using only one eval metric & Single metric gives an incomplete picture; different metrics reveal different failure modes. \\
\bottomrule
\end{tabular}
\caption{Failure modes definition.}
\label{tab:fm-taxonomy}
\end{table}

\subsection{Dataset Details}
\label{app:dataset-details}

\Cref{tab:dataset-characteristics} shows the characteristics of the tasks we curated, including source information, number of tasks by type, median row, and median of features. \Cref{tab:dataset-domains} shows the domain distribution.

\begin{table}[h]
\centering
\small
\begin{tabular}{@{}lrrrrrr@{}}
\toprule
\textbf{Source} & \textbf{N} & \textbf{Regr.} & \textbf{Bin.-cls.} & \textbf{Multi-cls.} & \textbf{Med. rows} & \textbf{Med. feat.} \\
\midrule
Kaggle & 44 & 18 & 17 & 8 & 100,820 & 24 \\
MLE-bench & 8 & 6 & 2 & 0 & 6,976 & 13 \\
OpenML & 36 & 10 & 15 & 11 & 3,104 & 18 \\
STRABLE & 108 & 76 & 13 & 19 & 7,710 & 17 \\
UCI & 24 & 12 & 7 & 5 & 11,805 & 13 \\
\midrule
\textbf{Total} & 220 & 122 & 54 & 43 & 8,693 & 18 \\
\bottomrule
\end{tabular}
\caption{Dataset characteristics across the 220 benchmark tasks.}
\label{tab:dataset-characteristics}
\end{table}

\begin{table}[h]
\centering
\begin{tabular}{@{}lr@{}}
\toprule
\textbf{Domain} & \textbf{Tasks} \\
\midrule
Healthcare & 50 \\
Finance & 47 \\
Science & 25 \\
Retail & 23 \\
Social & 21 \\
Energy & 17 \\
Transportation & 14 \\
Education & 11 \\
Environment & 8 \\
Mixed & 3 \\
Bioinformatics & 1 \\
\midrule
\textbf{Total} & 220 \\
\bottomrule
\end{tabular}
\caption{Domain distribution across the 220 benchmark tasks.}
\label{tab:dataset-domains}
\end{table}

\subsection{Human Annotator Agreement}
\label{app:human}
To assess the reliability of the LLM judge, a human annotator independently re-labeled all 15 failure modes on 19 randomly sampled pipelines (11 of the 30 originally sampled pipelines were excluded because their generated code was no longer recoverable on disk). Agreement was scored on a binary present/not-present basis per failure mode; a diagnostic check showed the annotator's absent/N/A distinction did not track the judge's own applicability gating consistently in either direction, so that distinction was not scored as disagreement. Across all 285 (pipeline × failure-mode) ratings, the judge agreed with the human annotator on 89.5\% of ratings (Cohen's $\kappa = 0.62$, 95\% CI [0.50, 0.73], via a pipeline-cluster bootstrap), indicating substantial agreement. Precision and recall for the present class were comparable (precision = 0.71, recall = 0.65), with no significant directional bias toward over- or under-flagging failures ($p = 0.29$). Disagreement concentrated in a small subset of failure modes, i.e., naive ensembling and reliance on a single evaluation metric (each at 0.68 agreement), while the remaining 13 failure modes achieved $\geq$0.79 agreement, 9 of them $\geq$0.89.

\begin{table}[t]
\centering
\small
\label{tab:validation-study-agreement}
\begin{tabularx}{\textwidth}{@{} l >{\raggedright\arraybackslash}X c c c c c @{}}
\toprule
\textbf{FM} & \textbf{Failure Mode} & \textbf{$n$} & \textbf{Agreement} & \textbf{Precision} & \textbf{Recall} & \textbf{Human Present} \\
\midrule
FM01 & Wrong task formulation$^\dagger$           & 19 & 0.95 & 0.00 & --   & 0 \\
FM02 & Peeking at test data during EDA$^\dagger$  & 19 & 0.95 & 0.00 & --   & 0 \\
FM03 & Not cleaning data                          & 19 & 0.95 & 0.80 & 1.00 & 4 \\
FM04 & Imputation before splitting                & 19 & 0.79 & 0.50 & 0.25 & 4 \\
FM05 & Ignoring temporal order$^\dagger$          & 19 & 0.95 & 0.50 & 1.00 & 1 \\
FM06 & No train/val split                         & 19 & 0.89 & 0.67 & 1.00 & 4 \\
FM07 & No feature engineering                     & 19 & 1.00 & 1.00 & 1.00 & 5 \\
FM08 & Scaling before splitting                   & 19 & 0.89 & 1.00 & 0.33 & 3 \\
FM09 & Feature selection before splitting$^\dagger$& 19 & 0.84 & 0.00 & 0.00 & 1 \\
FM10 & Wrong model type for task$^\dagger$        & 19 & 1.00 & --   & --   & 0 \\
FM11 & No regularisation in DL$^\dagger$          & 19 & 1.00 & --   & --   & 0 \\
FM12 & Naive ensembling                           & 19 & 0.68 & 0.50 & 0.67 & 6 \\
FM13 & Reusing test set across variants$^\dagger$ & 19 & 0.95 & --   & 0.00 & 1 \\
FM14 & Single eval of stochastic model            & 19 & 0.89 & 1.00 & 0.50 & 4 \\
FM15 & Using only one eval metric                 & 19 & 0.68 & 1.00 & 0.62 & 16 \\
\midrule
\multicolumn{2}{l}{\textbf{Overall} ($\kappa=0.62$)} & 285 & 0.89 & 0.71 & 0.65 & 49 \\
\bottomrule
\end{tabularx}
\caption{Human-vs-LLM-judge agreement on 19 randomly sampled, independently annotated pipelines. Scored as Present vs. Not-Present (all 15 failure modes per pipeline). Cohen's $\kappa=0.62$ (95\% CI [0.50, 0.73]) indicates moderate, above-chance agreement. Precision exceeds recall generally (judge-as-predictor, human as ground truth), indicating the judge is conservative rather than strict: reported failure rates in the main results are more likely an under- than an over-estimate.}
\end{table}

\subsection{Failure Analysis}

In \Cref{tab:gated-fm-applicable-counts} we show the number of applicable pipelines for each of the failure modes that require a specific pipeline stage or type of data to be present. 

\begin{table*}[h]
\centering
\small
\setlength{\tabcolsep}{5pt}
\begin{tabular}{ll rrrrrr}
\toprule
\textbf{Agent} & \textbf{Model} & \textbf{$n$ judged} &
\rotatebox{75}{\tiny Ignoring temporal order} &
\rotatebox{75}{\tiny No regularisation in DL} &
\rotatebox{75}{\tiny Reusing test set across variants} &
\rotatebox{75}{\tiny Single eval of stochastic model} &
\rotatebox{75}{\tiny Naive ensembling} \\
\midrule
\multirow{3}{*}{\textbf{ReAct}}
  & Gemini-3.7 & 219 & 137 & 0   & 0   & 211 & 198 \\
  & Qwen-4B    & 220 & 138 & 0   & 0   & 219 & 1   \\
  & Qwen-27B   & 220 & 138 & 0   & 124 & 217 & 118 \\
\midrule
\multirow{3}{*}{\textbf{AIDE}}
  & Gemini-3.7 & 219 & 137 & 3   & 41  & 215 & 215 \\
  & Qwen-4B    & 211 & 133 & 0   & 0   & 190 & 3   \\
  & Qwen-27B   & 185 & 112 & 0   & 4   & 176 & 27  \\
\midrule
\multirow{3}{*}{\textbf{MLEvolve}}
  & Gemini-3.7 & 219 & 124 & 134 & 118 & 194 & 189 \\
  & Qwen-4B    & 65  & 45  & 2   & 1   & 21  & 2   \\
  & Qwen-27B   & 203 & 125 & 45  & 46  & 194 & 98  \\
\midrule
\multicolumn{2}{l}{\textbf{Total}} & 1761 & 1089 & 184 & 334 & 1637 & 851 \\
\multicolumn{2}{l}{\textit{as \% of total}} & -- & 61.8\% & 10.4\% & 19.0\% & 93.0\% & 48.3\% \\
\bottomrule
\end{tabular}
\caption{Number of applicable pipelines per failure mode, by agent and model.}
\label{tab:gated-fm-applicable-counts}
\end{table*}

\begin{figure}[h]
    \centering
    \includegraphics[width=0.6\linewidth]{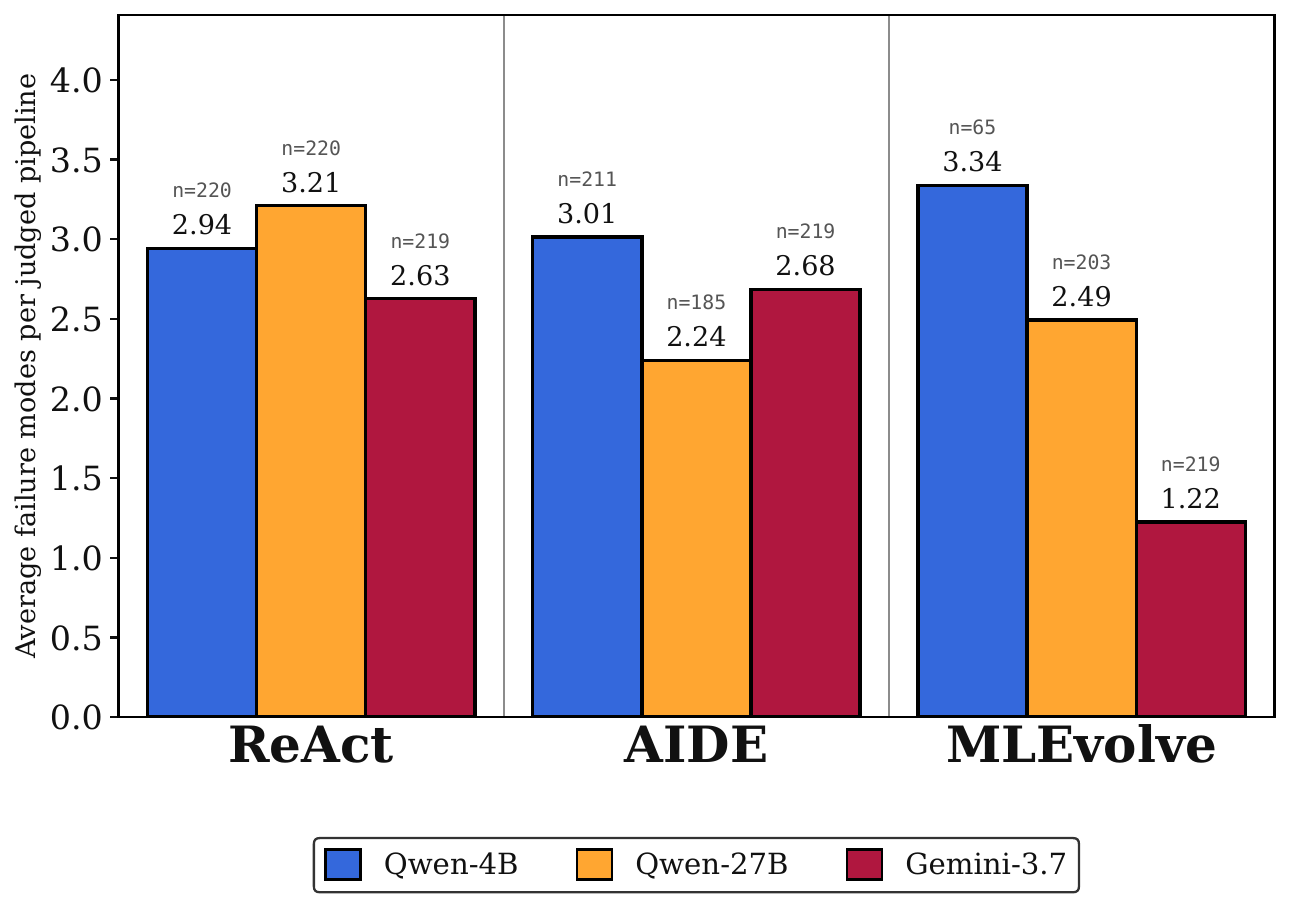}
    \caption{\textbf{Average failure modes per pipeline by agent-model configuration.}}
    \label{fig:avg_fm_agent}
\end{figure}

\begin{figure}[h]
    \centering
    \includegraphics[width=0.8\linewidth]{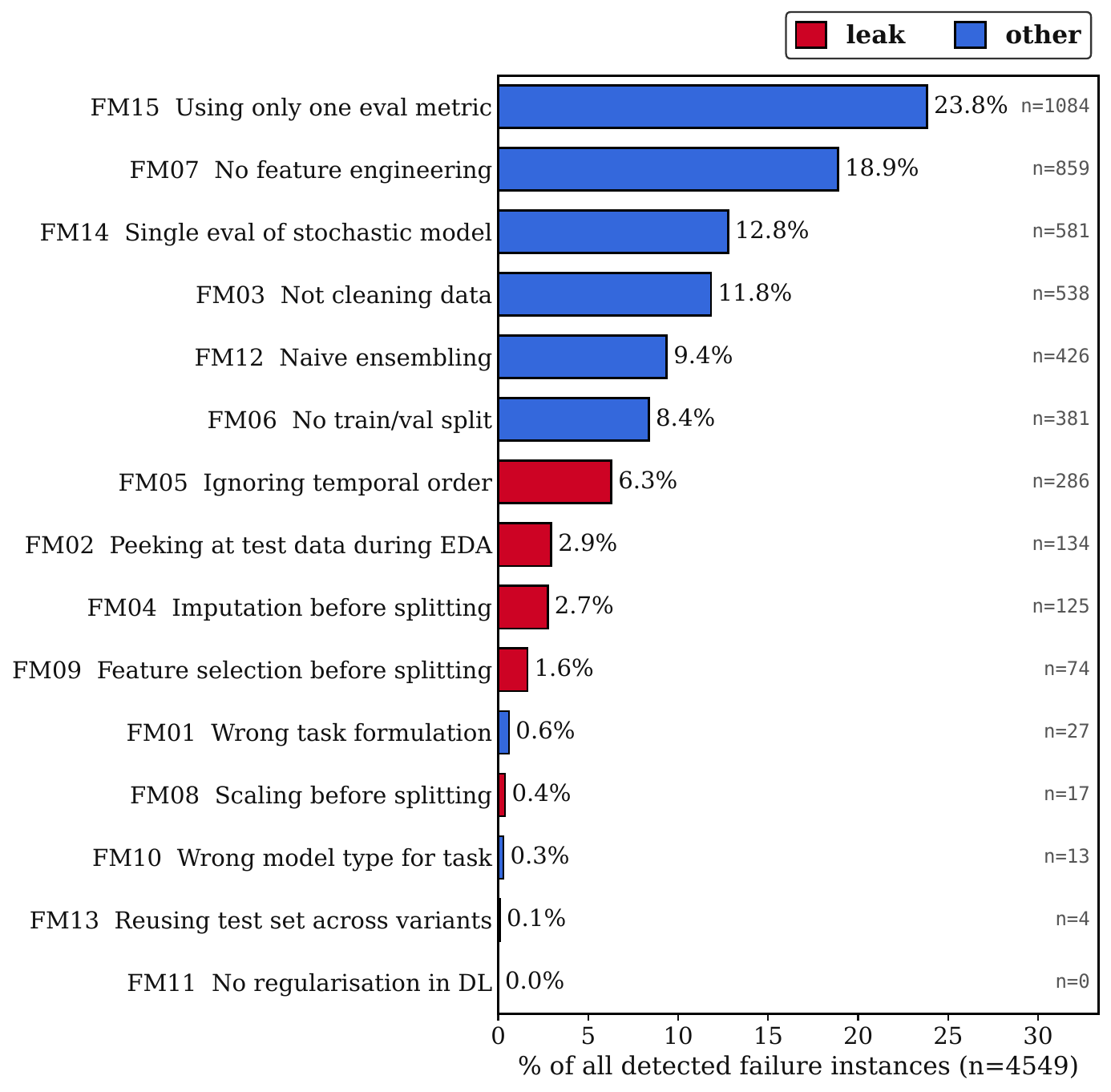}
    \caption{\textbf{Failure mode distribution. 4,557 failures detected over 1,761 pipelines, averaging 2.58 failures per pipeline.}}
    \label{fig:appen_fm_distribution}
\end{figure}

\subsection{Effects of failures on predictive performance}
\label{sec:residual-rank-details}

To calculate the failure modes' effect on the predictive performance, we provide the correlation between the two variables. However, scores alone are not comparable because tasks use different metrics. Therefore, we use the rank among all solutions as the measurement of predictive performance. 
We run generated solutions on the tasks and calculate the score on the pre-processed test set. Finally, we rank the solutions across all 9 configurations. 
A naive correlation between failure modes and predictive rank is confounded by model and agent capability. For example, \qwenSmall may exhibit fewer detected failures because it produces simpler pipelines, while also ranking poorly. We therefore estimate the effect of each failure mode after controlling for model and agent identity.

$$\text{rank} \sim \text{present} + C(\text{model}) + C(\text{agent})$$

For each failure mode, we fit a model that predicts a pipeline's rank from three things: whether the failure mode is present, which model produced it, and which agent produced it. Giving the model and agent their own separate adjustments, a model that's simply bad overall doesn't get mistaken for the failure mode itself being harmful. The one number we care about is how much presence of the failure mode shifts the rank, once that adjustment is made.

\begin{figure}[h]
    \centering
    \includegraphics[width=\linewidth]{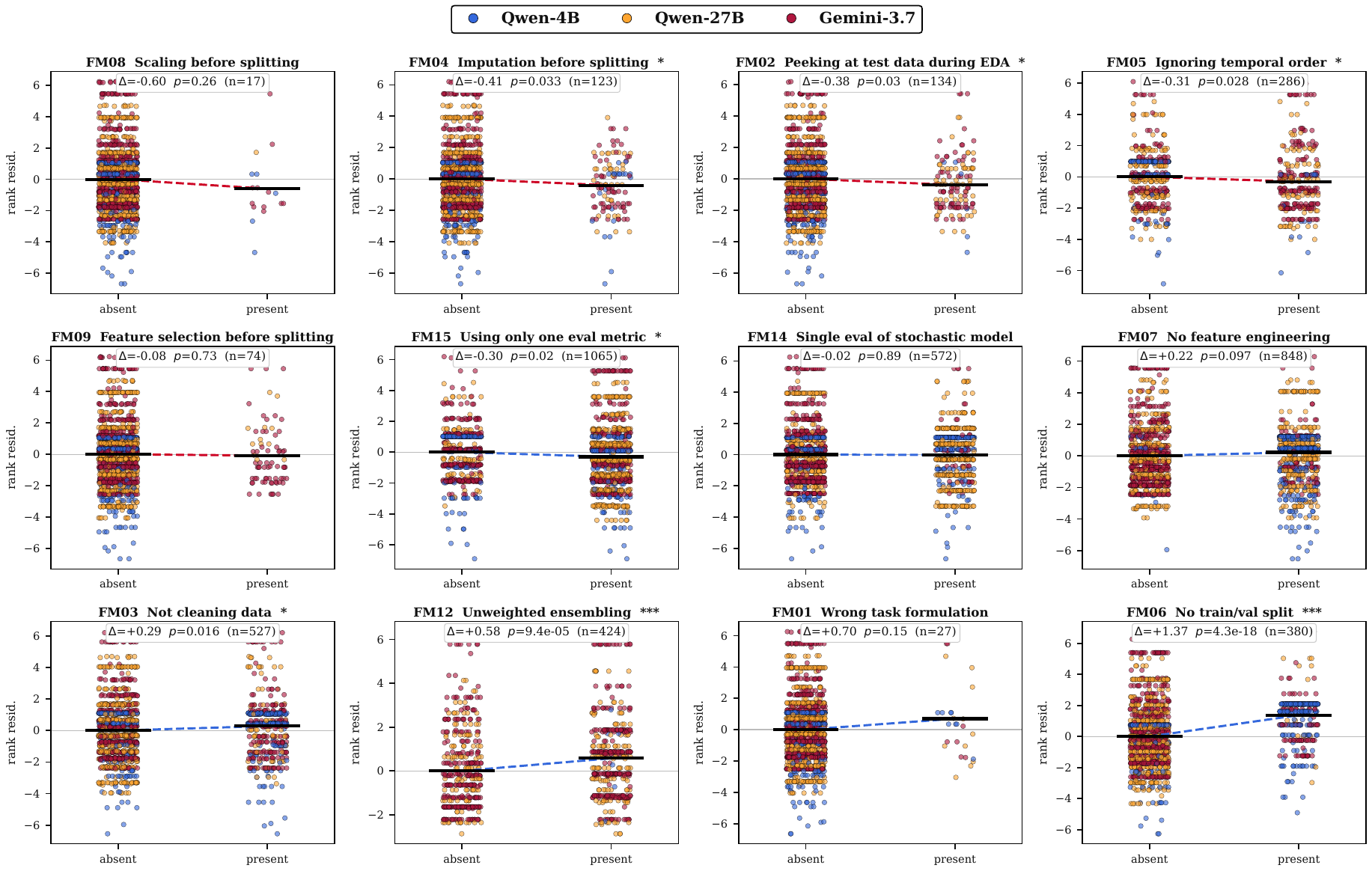}
    \caption{Effects of each failure mode on predictive performance measured in residual rank. }
    \label{fig:appen_res_rank}
\end{figure}

For visualization in Figure~\ref{fig:appen_res_rank}, we use the corresponding partial residual:
$$ r_i = \beta_1\text{present}_i + \hat{\varepsilon}_i. $$
Its group means are $0$ when the failure is absent and $\beta_1$ when present, so the plotted difference directly represents the adjusted effect.

\subsection{Failure Modes beyond the Taxonomy}
\label{app:fm_beyond_taxonomy}

In addition to the failure modes captured in the taxonomy, we evaluate the additional failure modes the LLM-judge reports as a list of free-text failures (Section~\ref{sec:llmag}). Across pipelines, the judge flags 2,294 additional failures. We filter these failures to only include actual violations of data science principles, filtering out other issues flagged by the judge, such as failures to produce code or implementation bugs, yielding 1,343 additional data science failures. We assign these failure modes into categories that we extract from manually screening them and then assign a category to each failure by independent keyword matching and through an LLM-judge (\qwen{3.5}{9b}), where disagreements are manually resolved. On a sample of 59 reports that were labelled blindly, the final labels agree in 95\% of the cases.

\textbf{Additional failure modes.} \Cref{tab:judge-other-fm-heatmap} shows the three most prominent categories. \textit{Validation$\rightarrow$train leakage} concerns instances where information from the validation data leaks into the trained model, without involving the test set. For instance, preprocessing or target encoding fitted on the full training data before cross-validation, early stopping on the fold the model is scored on, and ensemble weights fitted and scored on the same predictions. \textit{Test$\rightarrow$train leakage} encompasses leakage of test data into the training that falls outside of the taxonomy, such as transforms (e.g., label encoding, TF-IDF, or aggregated features) that are fitted on the training and test data. \textit{Flawed training protocol} includes training without early stopping, predicting with the model of a single fold, and not retraining the final model on the full training data. Less frequent additional categories are inconsistent preprocessing of training and test data, nominal features treated as continuous, identifier columns used as features, target leakage through a feature, and incorrectly implemented metrics, each occurring in at most 6\% of the pipelines.

\begin{table}[t]
\centering
\small
\setlength{\tabcolsep}{3pt}
\begin{tabular}{llccc}
\toprule
\textbf{Agent} & \textbf{Model} & \rotatebox{75}{\tiny Validation$\rightarrow$train leakage} & 
\rotatebox{75}{\tiny Test$\rightarrow$train leakage} & 
\rotatebox{75}{\tiny Flawed training protocol} \\
\midrule
\multirow{3}{*}{\textbf{ReAct}}  & Qwen-4B & \cellcolor[HTML]{FEEFD9}{\color[HTML]{000000}0.00} & \cellcolor[HTML]{FEEFD9}{\color[HTML]{000000}0.00} & \cellcolor[HTML]{FEE0B7}{\color[HTML]{000000}0.03} \\
 & Qwen-27B & \cellcolor[HTML]{FDCE98}{\color[HTML]{000000}0.08} & \cellcolor[HTML]{FEE4C0}{\color[HTML]{000000}0.02} & \cellcolor[HTML]{FDD09A}{\color[HTML]{000000}0.07} \\
 & Gemini-3.7 & \cellcolor[HTML]{FDB57E}{\color[HTML]{000000}0.15} & \cellcolor[HTML]{FDB57E}{\color[HTML]{000000}0.15} & \cellcolor[HTML]{FDC089}{\color[HTML]{000000}0.12} \\
\midrule
\multirow{3}{*}{\textbf{AIDE}}  & Qwen-4B & \cellcolor[HTML]{FDB27B}{\color[HTML]{000000}0.16} & \cellcolor[HTML]{FEEFD8}{\color[HTML]{000000}0.01} & \cellcolor[HTML]{F88154}{\color[HTML]{000000}0.29} \\
 & Qwen-27B & \cellcolor[HTML]{FDD39D}{\color[HTML]{000000}0.07} & \cellcolor[HTML]{FEE8C8}{\color[HTML]{000000}0.02} & \cellcolor[HTML]{FEE2BB}{\color[HTML]{000000}0.03} \\
 & Gemini-3.7 & \cellcolor[HTML]{FDBF88}{\color[HTML]{000000}0.13} & \cellcolor[HTML]{F2704C}{\color[HTML]{000000}0.35} & \cellcolor[HTML]{FEECD1}{\color[HTML]{000000}0.01} \\
\midrule
\multirow{3}{*}{\textbf{MLEvolve}}  & Qwen-4B & \cellcolor[HTML]{FDD9AA}{\color[HTML]{000000}0.05} & \cellcolor[HTML]{FFFFFF}{\color[HTML]{000000}0.00} & \cellcolor[HTML]{FDD9AA}{\color[HTML]{000000}0.05} \\
 & Qwen-27B & \cellcolor[HTML]{F4764F}{\color[HTML]{000000}0.33} & \cellcolor[HTML]{FEEFD8}{\color[HTML]{000000}0.00} & \cellcolor[HTML]{FEE5C3}{\color[HTML]{000000}0.02} \\
 & Gemini-3.7 & \cellcolor[HTML]{DF432D}{\color[HTML]{FFFFFF}0.50} & \cellcolor[HTML]{FDBA83}{\color[HTML]{000000}0.14} & \cellcolor[HTML]{FEECD1}{\color[HTML]{000000}0.01} \\
\bottomrule
\end{tabular}
\caption{Share of pipelines for which the judge reports the three most prominent failure modes beyond the taxonomy.}
\label{tab:judge-other-fm-heatmap}
\end{table}

\textbf{Observations.} The additional failures follow the same pattern described in Section~\ref{sec:large-scale-fm-analysis}. Validation$\rightarrow$train leakage increases with model capacity, from 8\% of the pipelines for \qwenSmall to 16\% for \qwenMedium and 26\% for \geminiFlash{3.7}. It is most frequent for MLEvolve (33\% with \qwenMedium and 50\% with \geminiFlash{3.7}), whose pipelines frequently combine cross-validation with target encoding, early stopping, and blending, and it almost never occurs when there is no validation step, as for ReAct with \qwenSmall. MLEvolve with \geminiFlash{3.7}, the configuration with the fewest failures within the taxonomy, has the highest share of pipelines with a failure falling outside the taxonomy (57\%), indicating that it insufficiently accounts for leakage. Since the judge reports these failures without being prompted for them specifically, the shares should be interpreted as lower bounds on their actual prevalence.

\subsection{Procedural comparisons between agentic solutions and expert solutions}
\label{app:case-study}
\begin{figure}
    \centering

    \includegraphics[width=1\linewidth]{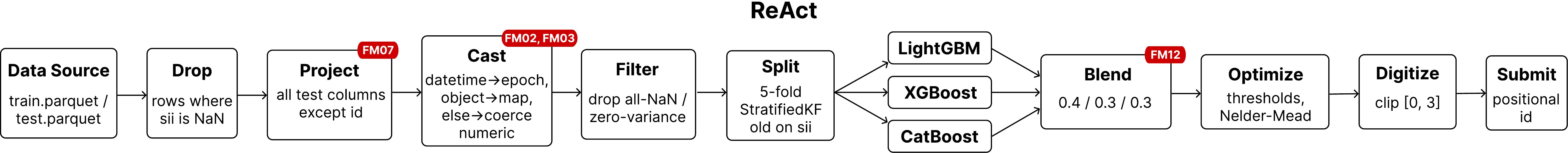}
    
    \includegraphics[width=.95\linewidth]{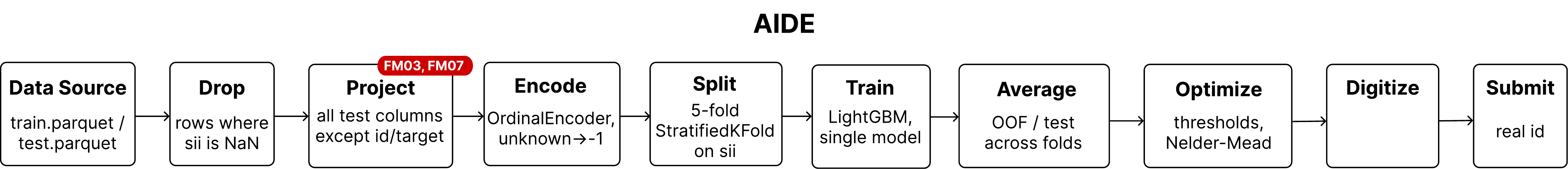}
    \vspace{.3em}
    
    \includegraphics[width=1\linewidth]{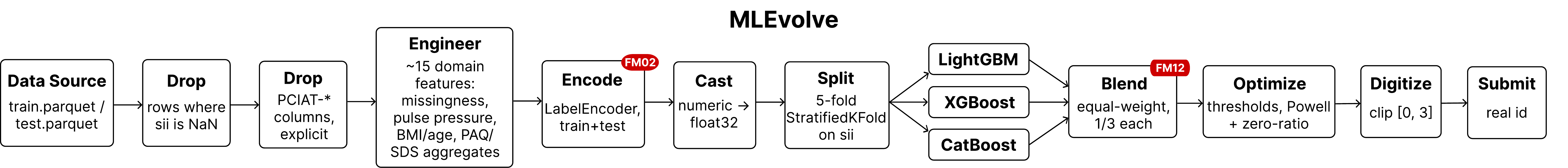}
    
    \includegraphics[width=1\linewidth]{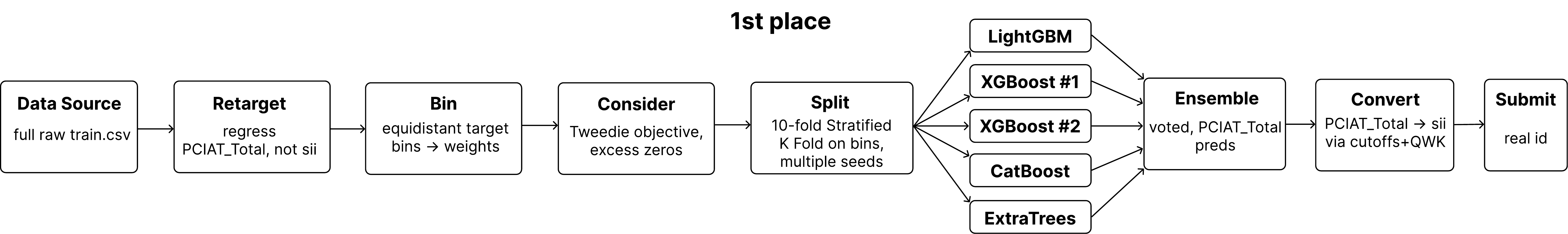}

    \vspace{-0.5em}
    \caption{Exemplar pipelines of DS agents generated solutions using \geminiFlash{3.7} v.s. human expert solution. 
    }
    \label{fig:combined-dags}
\end{figure}

In~\Cref{fig:combined-dags} we extract the key steps and procedures across solutions generated by three agents and expert solutions from Kaggle. 

\end{document}